%% file: main.tex
\documentclass[letterpaper, 10 pt, conference]{ieeeconf}  % Comment this line out if you need a4paper
\IEEEoverridecommandlockouts              
\usepackage{epsfig} % for postscript graphics files
\usepackage{fix-cm}
\usepackage{times} % assumes new font selection scheme installed
\usepackage{soul}
\usepackage{multirow}
\usepackage{multicol}
\usepackage{color}
\usepackage{comment}
\usepackage{amsmath}
\usepackage{wrapfig}
\usepackage{graphicx}
\usepackage[bookmarks=true]{hyperref}
\usepackage[ruled,vlined]{algorithm2e}
\usepackage{float}
\usepackage{subcaption}

\usepackage{bm}
\usepackage{amssymb}

\usepackage[T1]{fontenc}
\usepackage{tabularx} % Add this to your preamble
\usepackage{siunitx}  % Add this for better number formatting
\newcommand{\beq}{\begin{equation}}
\newcommand{\eeq}{\end{equation}}
\newcommand{\benum}{\begin{enumerate}}
\newcommand{\eenum}{\end{enumerate}}

\title{\LARGE \bf
TinySense: A Lighter Weight and More Power-efficient Avionics System for Flying Insect-scale Robots
}

\author{Zhitao Yu$^{*,1,3}$, Joshua Tran$^{*,2}$, Claire Li$^{2}$, Aaron Weber$^{1}$, Yash P.Talwekar$^{1}$, Sawyer Fuller$^{1,2}$
\thanks{This research was partially supported by the National Science Foundation through awards ECCS-2054850 and CNS-2235207 }% <-this % stops a space
\thanks{$^*$Authors contributed equally. }
\thanks{$^{1}$Department of Mechanical Engineering, University of Washington,
        Seattle, WA, USA}%
\thanks{$^{2}$Paul G. Allen School of Computer Science, University of Washington, Seattle, WA, USA}%
\thanks{$^{3}$Department of Applied Mathematics, University of Washington, Seattle, WA, USA}%
}
\begin{document}

\maketitle
\thispagestyle{empty}
\pagestyle{empty}

%%%%%%%%%%%%%%%%%%%%%%%%%%%%%%%%%%%%%%%%%%%%%%%%%%%%%%%%%%%%%%%%%%%%%%%%%%%%%%%%
\begin{abstract}

In this paper, we introduce advances in the sensor suite of an autonomous flying insect robot (FIR) weighing less than a gram. FIRs, because of their small weight and size, offer unparalleled advantages in terms of material cost and scalability. However, their size introduces considerable control challenges, notably high-speed dynamics, restricted power, and limited payload capacity. While there have been advancements in developing lightweight sensors, often drawing inspiration from biological systems, no sub-gram aircraft has been able to attain sustained hover without relying on feedback from external sensing such as a motion capture system. The lightest vehicle capable of sustained hovering---the first level of ``sensor autonomy''---is the much larger 28 g Crazyflie. Previous work reported a reduction in size of that vehicle's avionics suite to 187 mg and 21 mW. Here, we report a further reduction in mass and power to only 78.4 mg and 15 mW. We replaced the laser rangefinder with a lighter and more efficient pressure sensor, and built a smaller optic flow sensor around a global-shutter imaging chip. A Kalman Filter (KF) fuses these measurements to estimate the state variables that are needed to control hover: pitch angle, translational velocity, and altitude. Our system achieved performance comparable to that of the Crazyflie's estimator while in flight, with root mean squared errors of 1.573 deg, 0.186 m/s, and 0.136 m, respectively, relative to motion capture.  

\end{abstract}
%%%%%%%%%%%%%%%%%%%%%%%%%%%%%%%%%%%%%%%%%%%%%%%%%%%%%%%%%%%%%%%%%%%%%%%%%%%%%%%%
\input{1_introduction}
\input{2_methodology}
\input{3_sensor_suite_and_measurements}
\input{4_state_estimation}
\input{5_experiments}
\input{6_conclusion}

\addtolength{\textheight}{-0cm}  

\bibliographystyle{IEEEtran} 

\bibliography{references}

\end{document}

%% file: 1_introduction.tex
\section{INTRODUCTION}

Insect-scale robots have the potential to be deployed in the thousands or millions to perform ``fast, cheap, and out of control'' space missions, collective assembly tasks, or hazard detection, owing to their small size and low materials cost. Recent incarnations of sub-gram aircraft have demonstrated controlled flight~\cite{Ma2013,kim2022firefly}, as well as electromagnetically-mediated power delivery \cite{James2018, Jafferis2019, ozaki2021wireless}. But sensor autonomy, that is, the ability to hover without external feedback, has not yet been demonstrated below 10~g, never mind 1~g. A central question is  ``What are the minimum sensor suite and computation resources needed for the task of flight control?'' Of particular concern when designing such a control architecture is how the physics of scale affects sensing and control. The paper~\cite{fuller2022gyroscope} serves as a step toward a missing but complementary analysis to previous work that has largely focused on scaling effects on actuation and mechanics in small robots~\cite{Trimmer1989, caprari2001sugar, abbott2007robotics, paprotny2013small}. 

\begin{figure}
\centering\includegraphics[width=8cm]{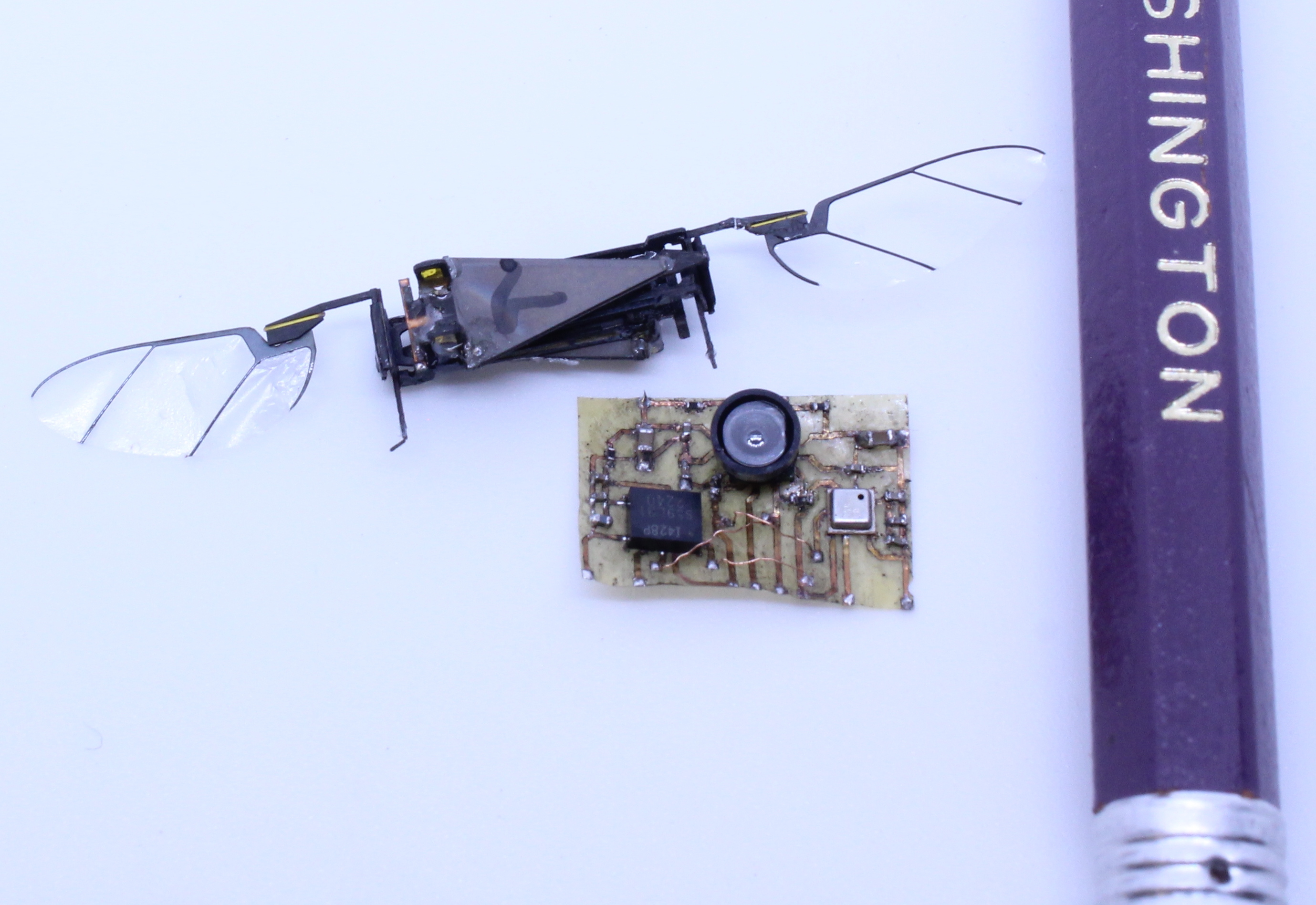}\caption{The TinySense sensor suite shown next to a U. Washington Robofly and a standard pencil for scale. The width of the sensor suite board is approximately 12~mm and its mass is 78.4~mg.\label{fig:robofly_and_gnat}\vspace{-0.6cm}}
\end{figure}

Insect-scale robots have extreme size, speed, weight, and power (SSWaP) constraints. ``Sensor autonomy'' including stable hovering, represents the first level of autonomy in the hierarchy proposed in~\cite{Floreano2015}, and is the foundation upon which higher level capability is built. We impose a requirement that sensing and computation take no more than 10\% of the power required to stay in the air; any more than this and flight time is severely impacted. This factor holds for important examples of autonomous drones, such as the 1.5~kg system in~\cite{Escobar-Alvarez2018} and the 30~g system in~\cite{Palossi2019}. This suggests a power budget of $\sim$20~mW for a 150~mg Robofly (Figure~\ref{fig:robofly_and_gnat}) that consumes 100~mW to fly (about 200~mW after losses from a 50\% efficient boost converter are factored in). 
% At a hundred-fold further reduction in mass to a 1~mg gnat robot, a hundred-fold reduction in power suggests a controller power budget of 200~µW. 

% \section{RELATED WORK}
%The sensing system is very crucial for robots. 
Large drones can carry larger sensor payloads such as light-based ranging systems (LIDAR) and the global positioning system (GPS). However, GPS does not work well in indoor environments~\cite{kjaergaard2010indoor} and small robots are not able to hold emissive sensors due to constraints in size, speed, weight, and power consumption (SSWaP). Recent work has introduced a 187~mg avoinics suite that consumes 21~mW~\cite{talwekar2022towards}, considered gyroscope-free avionics~\cite{fuller2022gyroscope}, and has investigated the effect of wing-induced vibrations~\cite{naveen2024hardware}. 

In this paper, we introduce a new avionics system for autonomous hovering flight that is even better tailored in mass and energy consumption for an insect-scale robot (Fig.~\ref{fig:sensors}) than previous work~\cite{talwekar2022towards}. To do so, we made two key refinements to lower power and mass. First, we replaced the power-hungry laser rangefinder with a pressure sensor. In recent years, these have attained sufficient precision to be used for the purpose of altitude estimation and control, even for insect-scale robots. Second, we replaced the optic flow sensor with a global shutter camera and computed optic flow on a microcontroller, resulting in a reduction in weight from 97~mg to 44~mg. As in~\cite{talwekar2022towards}, this new sensor suite is able to estimate the crucial variables that are needed for controlled flight: pitch angle, translational velocity, and altitude.  

Our contributions, in order of importance, are: 
\begin{enumerate}
    \item A reduction in weight of the sensor suite to 78.4~mg which is less than half of the previous lightest sensor suite~\cite{talwekar2022towards}. 
    \item Demonstration of accurate estimates of the state variables pitch angle, translational velocity, and altitude using data collected wirelessly from this sensor suite while mounted on a drone.
    \item Optic flow estimation on a microcontroller. 
\end{enumerate}

Subsequent sections elaborate on the dynamics, sensor configurations and measurement models, and validate our findings with data acquired wirelessly from our sensor suite, which is then analyzed on a desktop computer. 

%% file: 2_methodology.tex
\section{METHODOLOGY}
\subsection{Dynamics}
The basic dynamics of most small hovering devices are unstable in pitch angle and position. This was true for the first robotic flies~\cite{Perez-Arancibia2011} and EHD thruster actuated robots~\cite{Drew2017}, as well as larger aircraft such as the tailless Delfly Nimble~\cite{Karasek2018}. Techniques to stabilize them passively through modification of wing configuration~\cite{Fuller2019} or using air dampers~\cite{Fuller2017} are either unproven or result in undesirable steady-state oscillations, respectively. Our approach in this work is to first use linearized dynamics and observation models from which we can construct a Kalman observer of these unstable states so that they can be controlled. 

\begin{figure}
\centering\includegraphics[width=3.5cm]{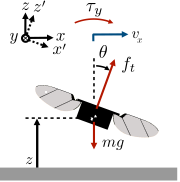}\caption{Forces and state variables of a small hovering Flying Insect Robot (FIR).} \label{fig:forces_and_velocities}
\vspace{-0.6cm}
\end{figure}
Fig.~\ref{fig:forces_and_velocities} depicts forces acting on
a flapping-wing robot such as a gnat robot or the Robofly, but the proposed model can apply to almost any small flying aircraft, including EHD-based thrusters~\cite{hari2020laser}. Near hover, the vehicle's dynamics are well approximated by a linear state-space model~\cite{Teoh2012,fuller2014controlling,Fuller2017} of the form $\dot{\boldsymbol q} = A\boldsymbol q + B \boldsymbol u + G\boldsymbol w$, where $\boldsymbol w$ is the process noise, assumed to be zero-mean Gaussian white noise with known covariance. Process noise $\boldsymbol w$ is included because it is a necessary consideration for the design of the estimator. 

To simplify the analysis, we restrict our attention to motion in the $x$-$z$ plane; the dynamics in the $y$-$z$ plane are very similar. We choose a minimal state vector $\boldsymbol{q}=[\theta,v_x,z]^{T}$, where $\theta$ is the pitch angle~(rad), $v_x$ is the vehicle's translational velocity in the world $x$-direction~(m/s), and $z$ is its vertical position or altitude~(m) as illustrated in Fig.~\ref{fig:forces_and_velocities}. When the nonlinear Euler-Lagrange equation dynamics are linearized by taking the Jacobian at $\theta = 0$, thrust force $f_{t}=mg$~(N), and $z=z_{d}$, where $z_d$ is the desired altitude~(m), the dynamics matrices are given by:
\begin{equation}
% \footnotesize
A = \left[\begin{matrix}0 & 0 & 0\\
g &  -\frac{b}{m} & 0\\
0 & 0 & 0
\end{matrix}\right],
\quad
B = \left[\begin{matrix}1\\
0\\
0
\end{matrix}\right],
\end{equation}

\noindent where $m$ is the vehicle's mass~(kg), and $b$ is the vehicle's translational drag proportionality constant~(Ns/m)~\cite{fuller2014controlling}. Note that because we are concerned in this paper about only estimation, we ignore inputs from the motor, and instead use the measurements of the gyroscope $\boldsymbol{u}=[\omega_m]$~(rad/s) as the only input. This is discussed further in Section~V.

%% file: 3_sensor_suite_and_measurements.tex
\section{Sensor Suite and Measurements}
Next we consider the sensor suite and state estimator. Our interest is in sensors that are either available commercially off-the-shelf, nearly so, or can be fabricated using the same tools that have been successful building insect robots such as the Robofly~\cite{Chukewad2021a} and Robobee~\cite{Wood2008, Ma2013}. A further constraint is that they require extremely low power for both the sensor itself and subsequent transduction and processing. This largely eliminates sensors that emit power, such as radar, sonar, depth cameras that emit structured light, and scanning laser rangefinders~\cite{Beyeler2009}, but permits passive sensing such as vision. The global positioning system is largely denied indoors and does not provide enough resolution anyway (1--10~m), and, further, requires significant power for signal processing. A lightweight and low-power sensing modality that can provide the necessary information to hover is  ``optic flow,'' which is defined as the velocity of the image (in rad/s) across the camera imager~\cite{Taylor2007, serres2016optic, serres2017optic}. In tandem with other sensors, optic flow can be used to estimate the distance to obstacles as the imager translates through space~\cite{Koenderink1987}. Methods for estimating optic flow include calculating luminance gradients locally~\cite{Lucas1981} (Lucas-Kanade), or image-wide~\cite{Horn1981}, as well as iterative search for matching blocks of pixels~\cite{Hartley2003}. Distances can also be estimated with a second camera and stereo vision using line matching~\cite{Wagter2014}, but this requires a second camera and a long ``baseline'' between cameras for accuracy, which adds excessive weight. 

\begin{figure}
\centering
\centering\includegraphics[width=8cm]{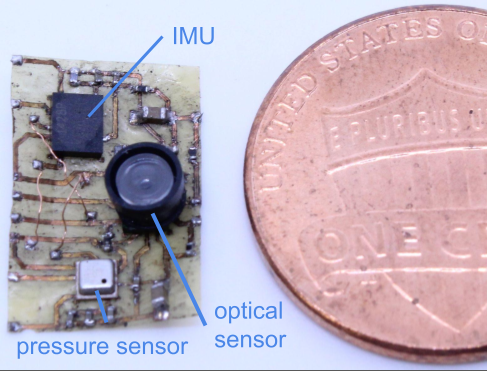}
\caption{Sensor suite (PAG7920LT optical sensor, ICM42688-P IMU, BMP388 pressure sensor) next to US 1 cent coin.}\label{fig:sensors}
\vspace{-0.6cm}
\end{figure}

Given these constraints, we propose the following set of onboard sensors: An optical sensor, a pressure sensor, and a gyroscope. We designed a customized microcircuit with the three sensors. (Fig.~\ref{fig:sensors}) 
\subsection{Camera and optic flow\label{sec:of}}
The system employs an optical sensor, the PAG7920LT (PixArt Imaging, Inc., Taiwan), to capture $160 \times 120$ QQVGA images with a $120^{\circ}$ field of view at 100~Hz over a 32~MHz SPI connection, with which the microcontroller computes optic flow via the Lucas-Kanade Method. The reason we select this camera is that, to our knowledge, it is the only global shutter camera that provides an SPI interface to transmit the image data. Other global shutter cameras require a special interface called MIPI (mobile industry processor interface), which is not available for any microcontroller. The majority of cameras employ a ``rolling shutter'' in which pixel rows are captured at sequential periods of time. A global shutter camera, by contrast, captures that all pixels at the same time, which ensures an accurate optic flow measurement.  

To minimize the data transmitted wirelessly over Bluetooth from our TinySense sensor suite, it is necessary to perform all optic flow calculations onboard. We wrote a Lucas-Kanade algorithm using the Arduino development environment in C++ to run on NRF52-based Adafruit Express board, chosen for its low size and weight.

The Lucas-Kanade algorithm is performed by calculating the following quantities:
 \[J = \left[ \begin{matrix}
I_x(p_1) & I_y(p_1) \\
I_x(p_2) & I_y(p_2) \\
... & ... \\
I_x(p_n) & I_y(p_n)
\end{matrix} \right],
\boldsymbol{v} = \left[ \begin{matrix}
V_x \\
V_y
\end{matrix} \right],
b = \left[ \begin{matrix}
-I_t(p_1) \\
-I_t(p_2) \\
... \\
-I_t(p_n)
\end{matrix} \right],
\]
\noindent where $p_1, p_2, \dots, p_n$ are the pixels in the image, $\boldsymbol{v}$ is the image flow in pixels per frame, and $I_x(p_i), I_y(p_i), I_t(p_i)$ are the image's horizontal, vertical, and temporal (time) partial derivatives respectively, at pixel $p_i$. Then, the optic flow is estimated (in pixels/frame) by taking the pseudoinverse $\boldsymbol{v} = (J^TJ)^{-1}J^Tb$.

To minimize computation for the limited microcontroller resources, we reduced the image resolution by a factor of 4 (producing a $40 \times 30$ resolution) by skipping pixels. 

The two most expensive operations involve computing $J^TJ$ and $J^Tb$. A $40 \times 30$ image produces an $J$ matrix with $(40-2) \times (30 - 2) = 1064$ rows and 2 columns and a $b$ matrix with 1064 rows and 1 column. Multiplying two matrices of size $k \times l$ and $l \times m$ requires at worst $kml$ multiplications. Thus, computing $J^TJ$ and $J^Tb$ require 4256 and 2128 multiplications, respectively. We neglect the additional operations required to compute their product (2 multiplications) and the $2 \times 2$ matrix inverse (6 multiplications and one divide), giving a total operations count of~6384.

We additionally implemented a form of sparse optic flow in which flow vectors were computed for twelve $10\times10$ patches, which are averaged to produce a final estimate. A $10 \times 10$ image produces an $J$ matrix with $(10-2) \times (10 - 2) = 64$ rows and 2 columns and a $b$ matrix with 64 rows and 1 column. Thus, computing $J^TJ$ and $J^Tb$ require approximately 256 and 128 multiplications, respectively. With twelve matrices of size $10 \times 10$, the total number of multiplication is $12 \times (256 + 128) = 4608$. This is a reduction of approximately $28\%$. 

To convert the optic flow from pixels/frame to angular velocity (rad/s), we found the calibration constant $\alpha=-0.027$~(rad/pixel) by rotating the TinySense by hand and performing a linear regression against its gyroscope readings.  

\subsection{Pressure sensor}

Altitude measurements were conducted by a Bosch BMP388 pressure sensor. This sensor estimates altitude as a proportion of the difference between the air pressure and sea level. We set it for 2$\times$ pressure oversampling and an IIR filter coefficient of 3 to maximize resolution. To compensate for variations in local pressure causing inaccurate altitude measurements, the sensor was assigned a bias equal to the average of the first 25 measurements where the sensor system was grounded; this bias was subtracted from every reading.

\subsection{Gyroscope}
For the onboard inertial measurement unit (IMU), we chose the TDK ICM42688-P because it is a recent model that is available in the smallest commercially-available package size of 2.5~$\times$~3~mm. It includes both a 3-axis gyroscope and a 3-axis accelerometer. In this work, we used only the gyroscope, which measures angular velocity through the detection of Coriolis forces within an electromechanical resonator. We configured the sensor to have a $\pm2000$ degrees/s full-scale (FS) range and enabled the sensor-provided low-noise mode to improve the sensitivity and minimize noise. During initial experiments, we noticed that the gyroscope's signal was noisier than expected and differed significantly from the Crazyflie's gyroscope, even after applying a digital low-pass filter. In handheld, non-flying experiments, however, the gyroscope proved accurate. We hypothesized that vibrations from the spinning propellers, which were at a higher frequency than the 100~Hz sampling rate, were introducing aliasing. We enabled the sensor's anti-aliasing low-pass filter with a cutoff frequency of 42~Hz, which is well below the Nyquist frequency associated with the 100~Hz sampling rate. We found that this setting significantly reduced the gyroscope's noise and error.

\begin{table}[!t]
    \caption{Comparison between TinySense and prior work}

\resizebox{\columnwidth}{!}{%
\begin{tabular}{ |c|c|c|c|c| } 
\hline
  & component & size  & mass & power \\
  &           & (mm)  & (mg) & (mW)\\
\hline
\multirow{5}{*}{prior work \cite{talwekar2022towards}} 
& IMU & 2.5$\times$3$\times$0.91 & 14 & 3 \\
& rangefinder & 4.9$\times$2.5$\times$1.56 & 16 & 5 \\
& optical flow & 5$\times$5$\times$3.08 & 97 & 12 \\
& board/discretes & -- & 60 & -- \\
& total & -- & 187 & 21 \\

\hline
\multirow{6}{*}{TinySense}
& IMU & 2.5$\times$3$\times$0.91 & 14 & 3 \\
& pressure & 2.0$\times$2.0$\times$0.75 & 14 & 2.6 \\
& camera chip & 2.2$\times$2.7$\times$6.9 & 24 & 10 \\
& lens & 3.8$\times$3.8$\times$1.95 & 20 & -- \\
& board/discretes & -- & 26.4 & -- \\
& total & -- & 78.4 & 15.6 \\
\hline

\end{tabular}
}
\vspace*{-16pt}
\end{table}

%% file: 4_state_estimation.tex
\section{State estimation}
To implement a Kalman Filter, we must first ascertain whether our system is observable at equilibrium. To show this, we linearize the observation model for the task of hovering. Note that the gyroscope provides a low-noise measurement of angular velocity $\boldsymbol \omega$. In cases like this in which the sensor provides more precise information than can be derived knowing the system's actuated inputs (e.g. torque applied by the aircraft's rotors or wings), it is common practice to reformulate the Kalman Filter to use the sensor's readings as ``input'' $\boldsymbol u$, and to ignore actual system input. This has an additional advantage that one less state variable is required for each such measurement. Our system includes gyroscope readings as one of its inputs and hence angular velocity $\omega$ does not appear among the state variables.  Our observation model for the suite of sensors is: 
\begin{equation}
\boldsymbol y=\left[\begin{matrix}
\Omega_m + n_o\\
z_m  + n_p\\

\end{matrix}\right],\label{eq:observationModel}
\end{equation} 
where $\Omega_m$ is the optic flow measurements~(rad/s) from optic flow sensor, $z_m$ is the measurement~(m) from the pressure sensor, and $n_o$ and $n_p$ are the noise terms for the two sensors. We assume noise terms are zero-mean Gaussian white noise, a standard assumption in sensor noise modeling to simplify the analysis. The measurement model for the optic flow in the $x$ direction, as measured by the optic flow camera is a nonlinear function of the state~\cite{fuller2022gyroscope}:
\begin{equation} 
\Omega_m = \omega_m - \frac{v_x}{z},\label{eq:of}
\end{equation}
where $\omega_m$ is the angular velocity measurement~(rad/s) from gyroscope. The optic flow measurement depends on $z$. We can take the Jacobian linearization at a desired altitude $z_d$:
\begin{equation} 
\begin{split}
\Omega_m(z) = \Omega_m(z_d) + O(z)
= \omega_m - \frac{v_x}{z_d} +  O(z).
\end{split}\label{eq:Jacobian}
\end{equation}

With this linearization, our measurement model can be cast into the Kalman Filter model, $\boldsymbol y = C\boldsymbol q +D \boldsymbol u + \boldsymbol n$, where
\begin{equation}
C=\left[\begin{matrix}
0 & -\frac{1}{z_d} & 0\\
0 & 0 & 1 
\end{matrix}\right],
\quad
D = \left[\begin{matrix}1\\
0
\end{matrix}\right].\label{eq:measurement}
\end{equation}

The observability matrix [$C$; $CA$] is full rank, meaning that a Kalman Filter will be able to estimate the system's state using the measurements given in Eq.~(\ref{eq:measurement}).

The Kalman Filter observer estimates $\hat{\boldsymbol{q}}$ by numerically integrating the system:
\begin{equation}
\dot{\hat{\boldsymbol{q}}}=A\hat{\boldsymbol{q}}+B\boldsymbol{u}+K(\boldsymbol{y}-C\hat{\boldsymbol{q}} - D\boldsymbol u),\label{eq:Kalman}
\end{equation} 
where $K$ is the Kalman gain. 

We calculated optic flow variance $R_{n1}$ to be 0.017 by subtracting its value from that predicted by mocap using Eq.~(\ref{eq:of}) during a constant-velocity horizontal translational flight. We calculated the variance of the pressure sensor $R_{n2}$ to be 0.0055 by taking measurements while holding it at a constant altitude for 60 seconds:
\begin{equation*}
R_n = \mathrm{diag}(R_{n1}, R_{n2}) = \mathrm{diag}(0.017, 0.0055).
\end{equation*}
To estimate the process noise covariance matrix $Q_n$, we employed a hybrid optimization approach, combining grid search for coarse parameter initialization, gradient descent for fine-tuning, and manual adjustment to minimize the RMSE between TinySense state estimates and mocap ground truth:
\begin{equation*}
Q_n = \mathrm{diag}(0.0124^2, 0.001^2, 0.22^2).
\end{equation*}
For the process noise (disturbance) matrix $G$, we assumed that the noise enters the system as white noise disturbance adding to the derivatives of the three states $\theta$, $v_x$, and  $z$. Based on these, the Kalman gain $K$ was computed using the \verb1lqe1 command in python-control~\cite{fuller2021python}:
\begin{equation*}
G = \left[\begin{matrix}
1 &  0 & 0\\
0 & 1 & 0 \\
0 & 0 & 1
\end{matrix}\right],
\quad
K = \left[\begin{matrix}
0.095 &  0\\
-1.32 & 0 \\
0 & 3
\end{matrix}\right].
\label{eq:G and K}
\end{equation*}
We implemented the Kalman Filter offline in Python by iteratively computing the state estimates $\hat{\boldsymbol{q}}_{i + 1}$ according to $
\hat{\boldsymbol{q}}_{i + 1} = \hat{\boldsymbol{q}}_{i} + \dot{\hat{\boldsymbol{q}}}_i \cdot dt$. 

% make sensor plots trial
\begin{figure*}[!t]
\centering
\centering\includegraphics[width=0.89\textwidth]{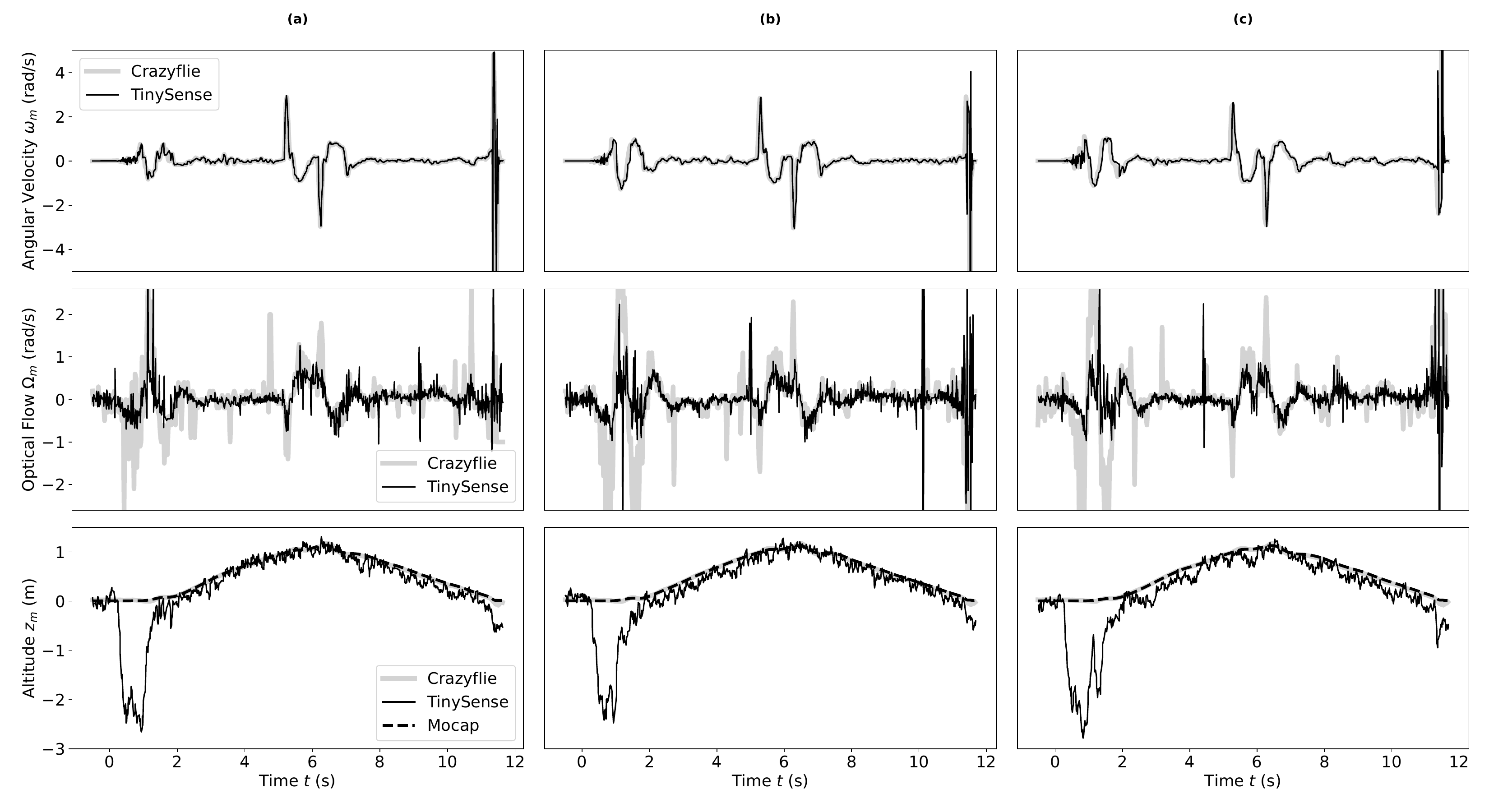}
\caption{Sensor measurements from the onboard gyroscope, optic flow sensor, and pressure sensor of TinySense are compared with those from the Crazyflie for three different flight experiments (a), (b), and (c); trajectories are described in the main text and corresponding state estimates in Fig.~\ref{state estimation}. In the altitude plots, altitude is measured by the Crazyflie's laser rangefinder, while the TinySense uses a pressure sensor; mocap data is also included for comparison. }\label{fig:sensor measurements}
\vspace{-16pt}
\end{figure*}

%% file: 5_experiments.tex
\section{Experiments}
\subsection{Robot platform}
We performed experiments using Crazyflie 2.0, a palm-sized quadrotor by Bitcraze (Sweden), in wind-free conditions in an indoor environment. In addition to its built-in gyroscope, the Crazyflie was equipped with the Flow deck v2 consisting of a downward-facing optic flow sensor and a laser rangefinder; it uses an Extended Kalman Filter to estimate its state in flight~\cite{Mueller2015}. 

\subsection{Motion capture system}
To provide a ground-truth comparison for the sensors, we used a twelve-camera OptiTrack Flex 13 motion capture arena, with spatial accuracy $\pm$ 0.20~mm. The Crazyflie drone was fitted with four retro-reflective motion capture markers and its position and orientation were recorded at 120~Hz. Velocity was estimated by taking the numerical derivative of the position data in post-processing.

\subsection{Experiment}
We connected our sensor suite to an NRF52840 Express (Adafruit), a board which supports data transmission through UART and 2.4~GHz Bluetooth Low Energy compatibility. We mounted the board as close to the geometrical center of the Crazyflie as possible to maintain a reasonable center of mass and minimize vibration interference to the IMU from the propellers. Power was supplied from the Crazyflie’s 250~mAh Lithium Polymer (LiPo) battery.

Each trial consisted of commanding the robot to rise to an altitude of 1~m, translate forward 1~m at 1~m/s, and then descend back to the ground. We ran this experiment three times. We collected measurements from the TinySense's sensors on a laptop as they were transmitted over Bluetooth at 100~Hz as binary data. Simultaneously, a second computer captured motion capture data at 120~Hz and used a Python-based script to capture sensor measurements and state estimates transmitted from the Crazyflie over Bluetooth at its maximum rate of 30~Hz. Each receiving device independently recorded a UTC timestamp, which was used to synchronize the data. The sensor measurements from the TinySense compared with the Crazyflie, and the motion capture (mocap) system's measurement for altitude are shown in Fig.~\ref{fig:sensor measurements}. 
\begin{figure*}[!t]
\centering
\centering\includegraphics[width=0.87\textwidth]{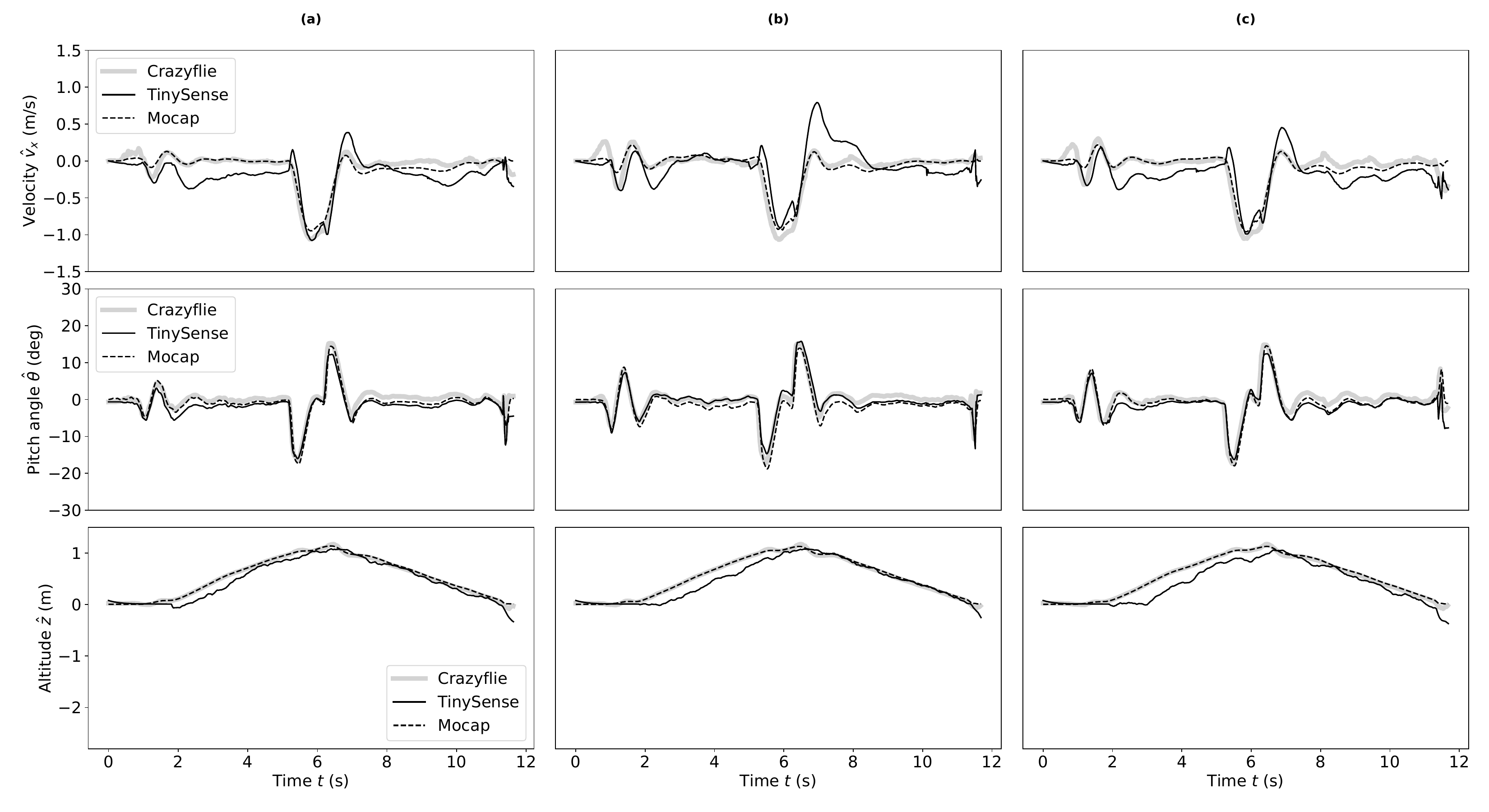}
\caption{Comparison between state estimates from TinySense's Kalman Filter, Crazyflie, and mocap. We conducted three flight experiments (a), (b), and (c), with the Kalman filter estimates starting at time zero. Pressure sensor measurements were ignored during the first 1.8~s because pressure changes due to the ground effect at takeoff confound readings.}\label{state estimation}
\vspace{-5pt}
\end{figure*}
{\setlength{\tabcolsep}{4pt}
\begin{table*}[!t]
  \centering
  \caption{Estimate of optic flow algorithm power use}
\begin{tabular}{|l|l|l|l|l|l|l|l|l|}
\hline
\shortstack{update \\ frequency (Hz)} & \shortstack{occurences/ \\ update} & task & \shortstack{single-cycle \\ operations} & \shortstack{int division \\ operations} & \shortstack{floating-point \\ division operations} & \shortstack{total \\ cycles/update} & \shortstack{total \\ cycles/s (Mhz)} & \shortstack{estimated \\ power (mW)} \\\hline
\multirow{3}{*}{100}  & 1                & skip frame & 8493                    & 0                        & 8                                   & \multirow{3}{*}{249781} & \multirow{3}{*}{25.0} & \multirow{3}{*}{5.5} \\ \cline{2-6}
                      & 12               & LK OF      & 18434                   & 128                      & 2                                   &                         &                              &                                 \\ \cline{2-6}
                      & 1                & copy data  & 1200                    & 0                        & 0                                   &                         &                              &                                 \\ \hline
\end{tabular}
\label{Table:of-power}
\\
\small
\vspace{-14pt}
\end{table*}
}
\subsection{Computational load}
Table~\ref{Table:of-power} provides an estimate of the power consumption of the optic flow algorithm, by calculating the number of clock cycles needed for each operation. Our estimate includes the matrix operations described in Section~\ref{sec:of}, as well as additional housekeeping operations, and is based on ARM Cortex-M4 instruction timings for addition, subtraction, multiplication, and memory read/write operations; integer divisions require 12 cycles at most and floating-point divisions require 14 cycles. Given that in our current system all sensors update at a rate of 100~Hz, we estimated that the total number of needed cycles per second is $\sim$25.0~Mhz. Operating the nRF52840 with an efficiency of 52~$\mu$A/MHz and a battery-powered voltage of 4.2~V, we estimate the power usage of the optic flow algorithm to be 5.5~mW.

\subsection{Results and discussion\label{sec:discussion}}
Fig.~\ref{state estimation} compares our sensor suite's state estimate to that of the Crazyflie's built-in estimator. States derived from the mocap system are considered to be ground truth. For the state estimator, time $t=0$ corresponds to when the gyroscope reading becomes significantly different from zero, indicating liftoff. Fig.~\ref{fig:sensor measurements} shows that a pressure increase occurs when the propellers first turn on, which can be observed as a drop in measured altitude. To mitigate unwanted effects of this in the Kalman Filter, we ignored (set to zero) altitude measurements $z_m$ during the first 1.8~s after liftoff. Nevertheless, once it is in the air, the altitude measurement matches quite closely with the laser rangefinder on the Crazyflie, though with slightly more noise.

Our system's estimate compares well with the Crazyflie's, despite receiving gyroscope readings at only 100~Hz, which is ten times slower~\cite{Mueller2015}. The key metric of pitch angle RMSE is approximately equal (Table~\ref{Table:RMSE}). Large errors in translational velocity $v_x$ appear around $t=5$ and $t=6$. These occur after sudden changes in attitude, during which angular velocity exceeds 3~rad/s (Fig.~\ref{fig:sensor measurements}). At these moments, we observed that TinySense optic flow readings are consistently lower than those of the Crazyflie. Lucas-Kanade cannot detect motion accurately if the displacement between frames is greater than a half pixel between image frames~\cite{barron1992performance}. This limits optic flow measurement to $\Omega_m<0.5 \alpha F = 1.4$~rad/s, where $\alpha$=0.027 is the angular increment between  pixels and $F$=100 is the frame rate (Hz). We believe that the lateral velocity estimate diverges during these fast maneuvers because it depends on erroneous optic flow  readings.

\begin{table}[!t]
    \caption{Root Mean Squared Error (RMSE) of estimates relative to mocap system} 

\resizebox{\columnwidth}{!}{%
\begin{tabular}{ |c|c|c|c| } 
\hline
     &  velocity $v_x$ & pitch angle $\theta$ & altitude $z$ \\
     & (m/s) & (deg) & (m) \\
\hline
Crazyflie & 0.075 $\pm$ 0.009 & 1.619 $\pm$ 0.267 & 0.021 $\pm$ 0.001 \\
\hline
TinySense & 0.186 $\pm$ 0.015 & 1.573 $\pm$ 0.166 & 0.136 $\pm$ 0.026 \\
\hline

\end{tabular} \label{Table:RMSE}
}
% \vspace*{-10pt}
\vspace{- 0.6 cm}
\end{table}

%% file: 6_conclusion.tex
\section{Conclusion}
In the present study, we introduce a new sensor suite with dramatically reduced mass, and associated optic flow and state estimation software that is compatible with the computational constraints of an onboard microcontroller small enough to fly aboard flying insect robots (FIRs) weighing less than a gram. Our proposed avionics suite diverges in two important ways from the 28~g Crazyflie helicopter, the lightest vehicle yet to perform controlled hovering, and previous work~\cite{talwekar2022towards}. First, we replaced a power-hungry laser rangefinder with a much more efficient (and slightly less ideal) and lighter pressure sensor. Second, we replaced an off-the-shelf optic flow sensor, which is available only in relatively heavy packages, with a custom camera with a global shutter and a custom-written Lucas-Kanade-based optic flow algorithm running onboard a 10~mg microcontroller small enough to fly onboard an FIR. We estimate the compute power usage is about 5~mW. Our system uses a linear Kalman Filter to estimate pitch angle, translational velocity, and altitude. This can be readily adapted to a nonlinear extended Kalman Filter for more generality~\cite{talwekar2022towards}. 

The cumulative weight of this sensor suite is only 78.4~mg, rather less than the 187~mg of previous work~\cite{talwekar2022towards}. With the addition of a 30~mg microcontroller and crystal, the overall system remains well within the 252~mg payload limit of the 143~mg robotic platform reported in~\cite{Fuller2019}. Our system demonstrates average RMSE values of 0.186~m/s for velocity, 1.573~deg for pitch angle, and 0.136~m for altitude when compared to the motion capture system. In comparison, the Crazyflie platform achieves RMSE values of 0.075~m/s, 1.619~deg, and 0.021~m for the same states during similar flight maneuvers. This shows that our system performs closely to the state-of-the-art Crazyflie in these scenarios. The pressure sensor showed negligible drift over a 60-second period, providing a usable altitude estimate as long as its output during takeoff is ignored. Future improvements include mitigating the lateral velocity errors that occur during rapid rotations that exceed the limit of Lucas-Kanade estimation~(Section~\ref{sec:discussion}) by increasing the frame rate or image blurring. Additionally, our patch-based optic flow estimation (Section~\ref{sec:of}) may allow outliers to be rejected based on their eigenvalues to eliminate low-contrast patches, reducing noise~\cite{barron1992performance}.

Future work will integrate TinySense along with a microcontroller and power hardware onto insect-scale robots such as the Robofly~\cite{james2018liftoff} to perform visual navigation~\cite{yu2022visual}. The Robofly performs faster accelerations than the Crazyflie and is a different vibration environment, but is otherwise similar. Previous work shows vibrations do not confound gyroscope readings~\cite{Fuller2014b, naveen2024hardware}. Algorithmic changes in optic flow may be needed to accommodate vibration, but our choice of a global shutter camera will mitigate tearing artifacts.